\documentclass{article}
\usepackage{frExamplee}
\usepackage{graphicx}
\usepackage{apalike}
\usepackage{setspace}

\usepackage{amsmath}
\usepackage{subcaption}

\title{Towards A COLREGs Compliant Autonomous Surface Vessel in a Constrained Channel}

\author{
James Connor Meyers\\
Field Robotics Laboratory, Northeastern University\\
United States Coast Guard\\
\And
Thomas Sayre McCord\\
MIT - WHOI Joint Program\\
Lilium GmbH\\
\AND
Zhiyong Zhang\\
Field Robotics Laboratory, Northeastern University\\
\And
Hanumant Singh\\
Field Robotics Laboratory, Northeastern University\\
\\
}

\begin{document}

\maketitle

\begin{abstract}
In this paper, we look at the role of autonomous navigation in the maritime domain. Specifically, we examine how an Autonomous Surface Vessel(ASV) can achieve obstacle avoidance based on the Convention on the International Regulations for Preventing Collisions at Sea (1972), or COLREGs, in real-world environments. Our ASV is equipped with a broadband marine radar, an Inertial Navigation System (INS), and uses official Electronic Navigational Charts (ENC). These sensors are used to provide situational awareness and, in series of well-defined steps, we can exclude land objects from the radar data, extract tracks associated with moving vessels within range of the radar, and then use a Kalman Filter to track and predict the motion of other moving vessels in the vicinity. A Constant Velocity model for the Kalman Filter allows us to solve the data association to build a consistent model between successive radar scans. We account for multiple COLREGs situations based on the predicted relative motion. Finally, an efficient path planning algorithm is presented to find a path and publish waypoints to perform real-time COLREGs compliant autonomous navigation within highly constrained environments. We demonstrate the results of our framework with operational results collected over the course of a 3.4 nautical mile mission on the Charles River in Boston in which the ASV encountered and successfully navigated multiple scenarios and encounters with other moving vessels at close quarters. 
\end{abstract}

\section{Introduction}

\subsection{Navigation Rules (COLREGs)}
The International Maritime Organization (IMO) promulgated the International Regulations for Preventing Collisions at Sea (COLREGs)\cite{COLREG_Convention_on_the_International_Regulations_for_Preventing_Collisions_at_Sea} in 1972. These are considered the "rules of the road" or the navigation rules to be followed by ships and other vessels at sea to prevent collisions between two or more vessels.

\begin{figure} [h]
    \centering
    \includegraphics[height=2.0in]{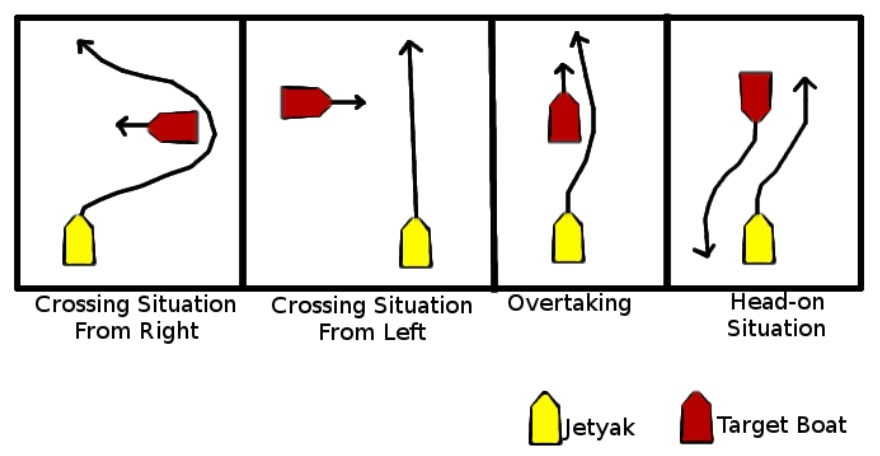}
    \caption{COLREGs maneuvers. The two interacting vessels are both required to follow COLREGs. We present the four fundamental scenarios, which encompass crossing, overtaking, and head-on situations between two ASVs. In covering these situations, our ASV (the Jetyak) is shown in yellow.}
    \label{COLREGs_maneuvers}
\end{figure}

One way to attempt safe navigation for USVs is to emulate human behavior as closely as possible through COLREGs. In this paper, we consider the situations represented in Figure \ref{COLREGs_maneuvers}. These represent the responsibilities of power-driven vessels in sight of each other to prevent collision while passing close to each other. A far more detailed description of responsibilities between vessels is provided in \cite{COLREG_Convention_on_the_International_Regulations_for_Preventing_Collisions_at_Sea}.

\subsection{Current Solutions}
While many solutions exist for COLREGs compliant autonomous surface vessels, none of these solutions use open-source hardware and software with low-cost sensors to perform COLREGs avoidance for both stationary and dynamic obstacles in congested waterways. 

There are expensive military applications such as the Common Unmanned Surface Vehicle (CUSV) created by the US Navy that operate under dynamic environments, but the research is highly proprietary, and the sensor suites utilized are extremely expensive. Other methods use lower-cost sensors and vessels, and some of them perform well. Blaich et al. \cite{Mission_integrated_collision_avoidance_for_USVs_using_laser_range_finder} propose a collision avoidance method without COLREGs constraints. Benjamin et al. \cite{A_method_for_protocol_based_collision_avoidance_between_autonomous_marine_surface_craft} follow COLREGs by sharing the action and position of multiple ASVs. Liu et al. \cite{Path_planning_algorithm_for_unmanned_surface_vehicle_formations_in_a_practical_maritime_environment} simulate a path planning algorithm in static environment. Dubey et al. \cite{VORRT_COLREGs_A_Hybrid_Velocity_Obstacles_and_RRT_Based_COLREGs_Compliant_Path_Planner_for_Autonomous_Surface_Vessels} and Naeem et al. \cite{COLREGs_based_collision_avoidance_strategies_for_unmanned_surface_vehicles} simulate the collision avoidance strategies based on COLREGs. Benjamin et al. \cite{COLREGS_based_navigation_of_autonomous_marine_vehicles} address the problem of safe navigation under COLREGs in open waters. Johansen et al. \cite{Ship_Collision_Avoidance_and_COLREGS_Compliance_Using_Simulation_Based_Control_Behavior_Selection_With_Predictive_Hazard_Assessment} and Zhang et al. \cite{A_distributed_anti_collision_decision_support_formulation_in_multi_ship_encounter_situations_under_COLREGs} outline methods for avoiding multiple dynamic obstacles in simulation. Kuwata et al. \cite{Safe_Maritime_Autonomous_Navigation_With_COLREGS_Using_Velocity_Obstacles} provide a velocity obstacles based solution for open water.

\subsection{Our Approach}
The method proposed in this paper tackles all these issues. By using the WHOI Jetyak \cite{The_WHOI_Jetyak_An_autonomous_surface_vehicle_for_oceanographic_research_in_shallow_or_dangerous_waters} and open-source hardware and software, costs are kept low compared to Naval and professional-grade sensor suites without significantly degraded performance. By using freely available marine charts to generate a binary occupancy grid of the environment, the Jetyak can avoid other vessels following COLREGs and be able to do so in an environment constrained by land areas, shoal water, and other navigational hazards. All the results presented in this paper are based on real data collected during with real vessels on the Charles River in Boston.

\begin{figure} [h]
    \centering
    \includegraphics[height=4.5in]{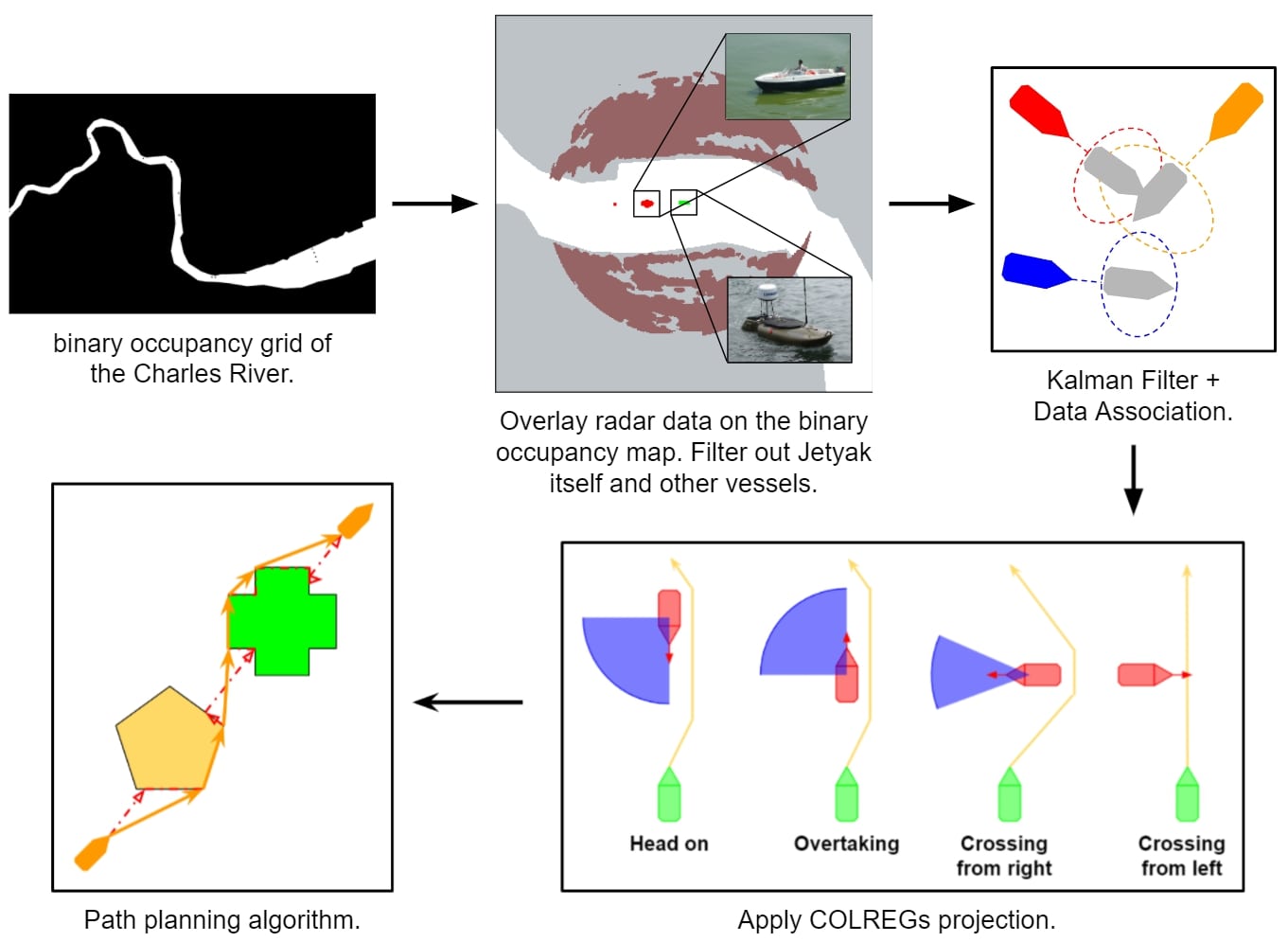}
    \caption{The Systems Level Diagram for our Methodology: A binary occupancy grid is created from an Electronic Navigation Chart (top left). As the ASV traverses the area it overlays radar data on the chart (top center). A Kalman filter and Data Association are used to determine target trajectories (top right). COLREGs considerations are then applied to steer the ASV (bottom right) as well as to replan the trajectory of the vessel ((bottom left).}
    \label{diagram}
\end{figure}

\section{The Jetyak ASV and its Sensor Suite}

The WHOI Jetyak was used as the ASV for tests presented in this paper, see  \cite{The_WHOI_Jetyak_An_autonomous_surface_vehicle_for_oceanographic_research_in_shallow_or_dangerous_waters} for more details. A Navico BR24 Radar was mounted on top of the Jetyak providing environmental sensing of dynamic and stationary obstacles, while a Vectornav VN-300 dual antennal internal navigation system was used to determine the Jetyak's own position and yaw.
The data acquisition is accomplished with ROS, and the use of the BR24 library \cite{br24} to collect radar data, MavRos \cite{mavros} to collect odometry information including speed, and to broadcast MavLink \cite{mavlink} messages including waypoints and vehicle information, and the Vectornav library \cite{vectornav} to collect heading and GPS information from the VN-300. ROS messages received include approximately 800 radar scanlines per second. Each scanline contains 512 ranges with intensity values from 0-255. The scan rate of the radar is one full rotation every 2.5 seconds (one full 360-degree scan). Highly accurate heading information is delivered at 100Hz from the VN-300 INS with errors of the order of approximately 0.2 degrees RMS. All Mavlink messages are updated at 10Hz, with GPS updates at $\sim$2Hz.

Marine charts are required for our algorithm, and they typically provide accurate descriptions of the land, water depths, navigational aids, and features useful to safe nautical navigation. The Electronic Navigation Chart (ENC) provides a good description of the large-scale static environment. In our work, ENCs are used to overlay radar data and perform avoidance of the shoreline, other navigational hazards, and ultimately avoid dynamic and stationary obstacles. Where NOAA does not provide ENCs, as in our case on the Charles river, open-source QGIS software \cite{QGIS} was used to create approximations of the land area using satellite imagery.

\section{Data Preprocessing}

\subsection{Binary Occupancy Grid}
\begin{figure} [h]
    \centering
    \includegraphics[height=2.5in]{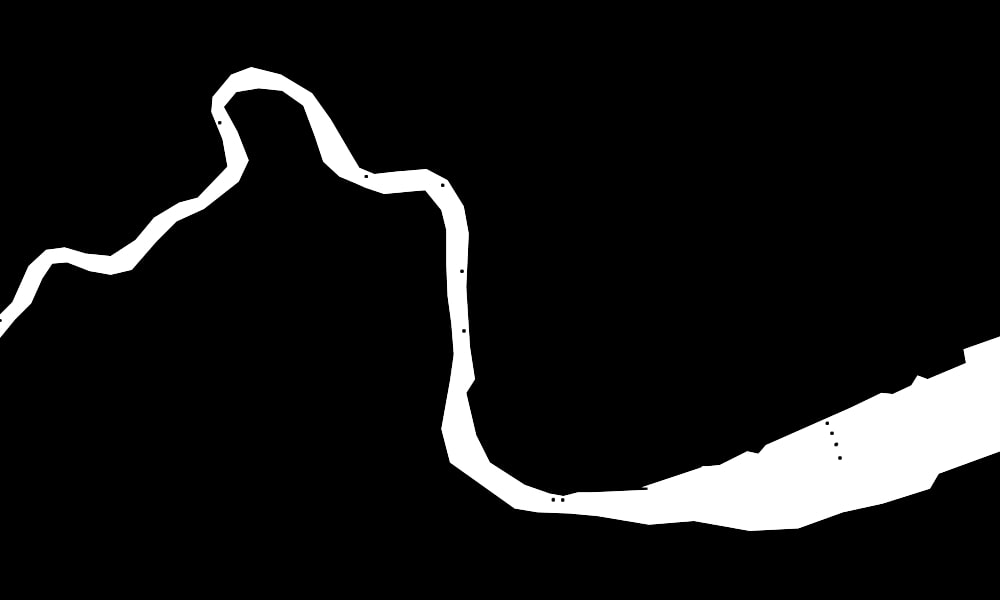}
    \caption{A binary occupancy grid of part of the Charles River in Boston generated by the QGIS. The black areas are land, while the white area corresponds to the river}
    \label{CharlesRiverbinaryoccupancygrid}
\end{figure}

A binary occupancy grid is used to represent space occupied by known navigational hazards such coastlines and bridges as well as target vessels. Any black areas are considered static obstacles on the map, and any white area is considered open water space. A path planner can determine an obstacle-free path based on this map. See Fig. \ref{CharlesRiverbinaryoccupancygrid} for a representation of the Charles River binary occupancy grid.

\subsection{Radar Processing}
\begin{figure}	
	\centering
	\begin{subfigure}[h]{3.0in}
		\centering
		\includegraphics[width=3.0in]{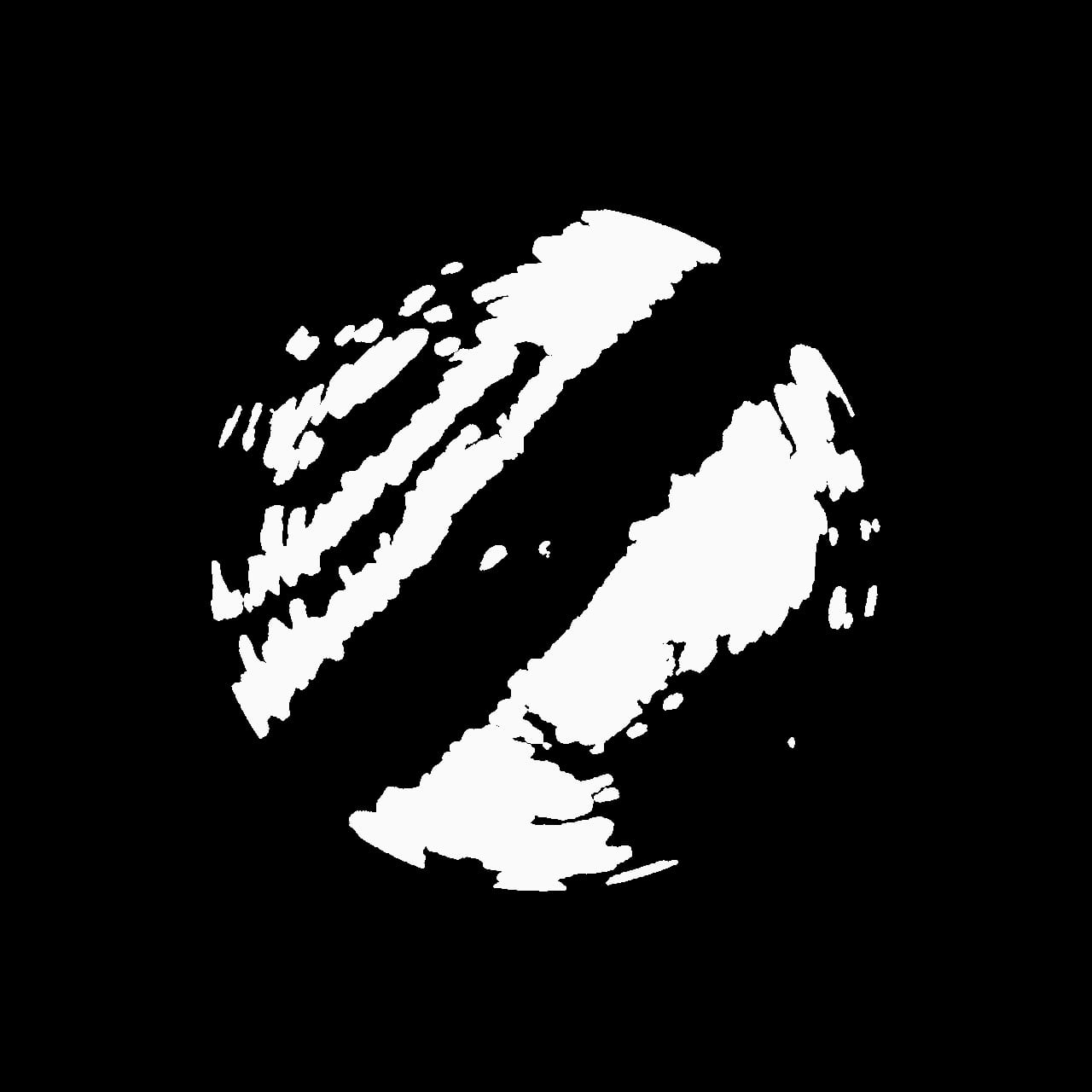}
		\caption{Raw radar data}		
	\end{subfigure}
	\hspace{.2in}
	\begin{subfigure}[h]{3.0in}
		\centering
		\includegraphics[width=3.0in]{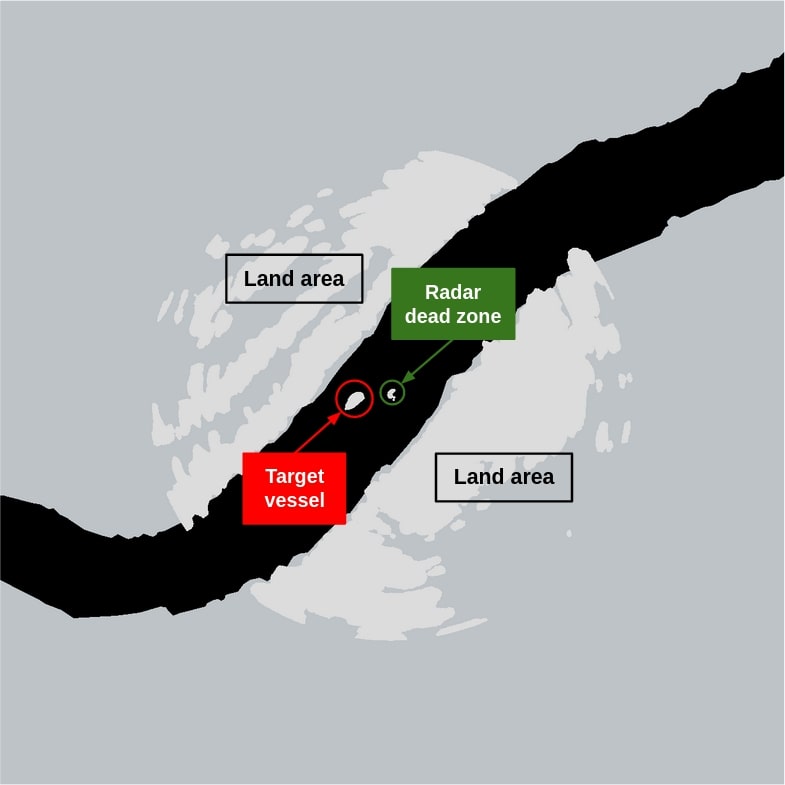}
		\caption{Radar data overlay}
	\end{subfigure}
	\caption{Overlaying the radar data on the binary occupancy grid of the Charles River.}
	\label{radar_data_overlay}
\end{figure}

We overlay the radar data on the binary occupancy grid map of the environment. It is completed by offsetting the radar scanline angle from the Jetyak's heading and global position. The global position data is computed in Universal Transverse Mercator coordinates (UTM). After data overlay, we can extract target vessels on the river and exclude land area in the next step as illustrated in Figure \ref{radar_data_overlay}.

\begin{figure} [h]
    \centering
    \includegraphics[height=2.0in]{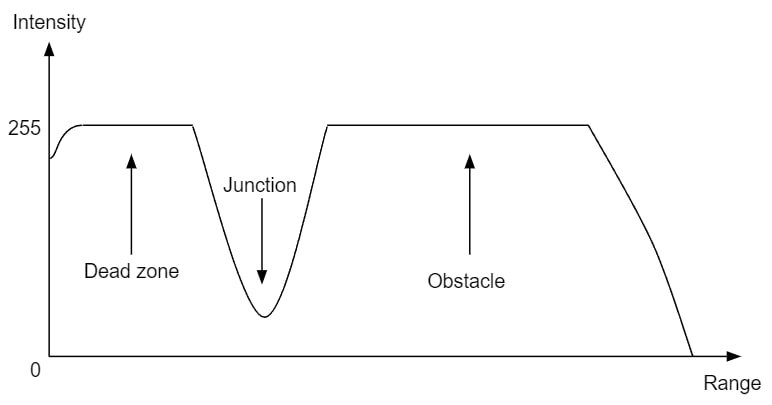}
    \caption{Excluding the dead zone. We find the lowest intensity point at the junction of the radar intensities and remove all the occupied grid cells from the start of the transmission to the junction point.}
    \label{exclude_dead_zone}
\end{figure}

There is a dead zone of returns near the radar, which appear as occupied grids but in reality are an artefact of the radar. The dead zone occupied grids can sometimes connect to real obstacles near the Jetyak. To differentiate between these two cases, we find the lowest intensity point at the junction of the radar intensities and radar signal returns before the junction point are considered the dead zone. Other signals after the junction are treated as obstacles. We filter out all the signals that are considered to be in the dead zone. See Fig. \ref{exclude_dead_zone}.

\subsection{Target Extraction}

As the first step of target tracking, the components of the radar scan corresponding to known land in the ENC map is removed.
In contrast to Blaich et al. \cite{Collision_Avoidance_for_Vessels_using_a_Low_Cost_Radar_Sensor} which removes objects based on whether the center point lies on land or water, we use a polygon based approach. As an offline processing step the ENC (or map from QGIS for the Charles River) is converted to polygons for fast online processing.

\begin{figure}	
	\centering
	\begin{subfigure}[h]{3.0in}
		\centering
		\includegraphics[width=3.0in]{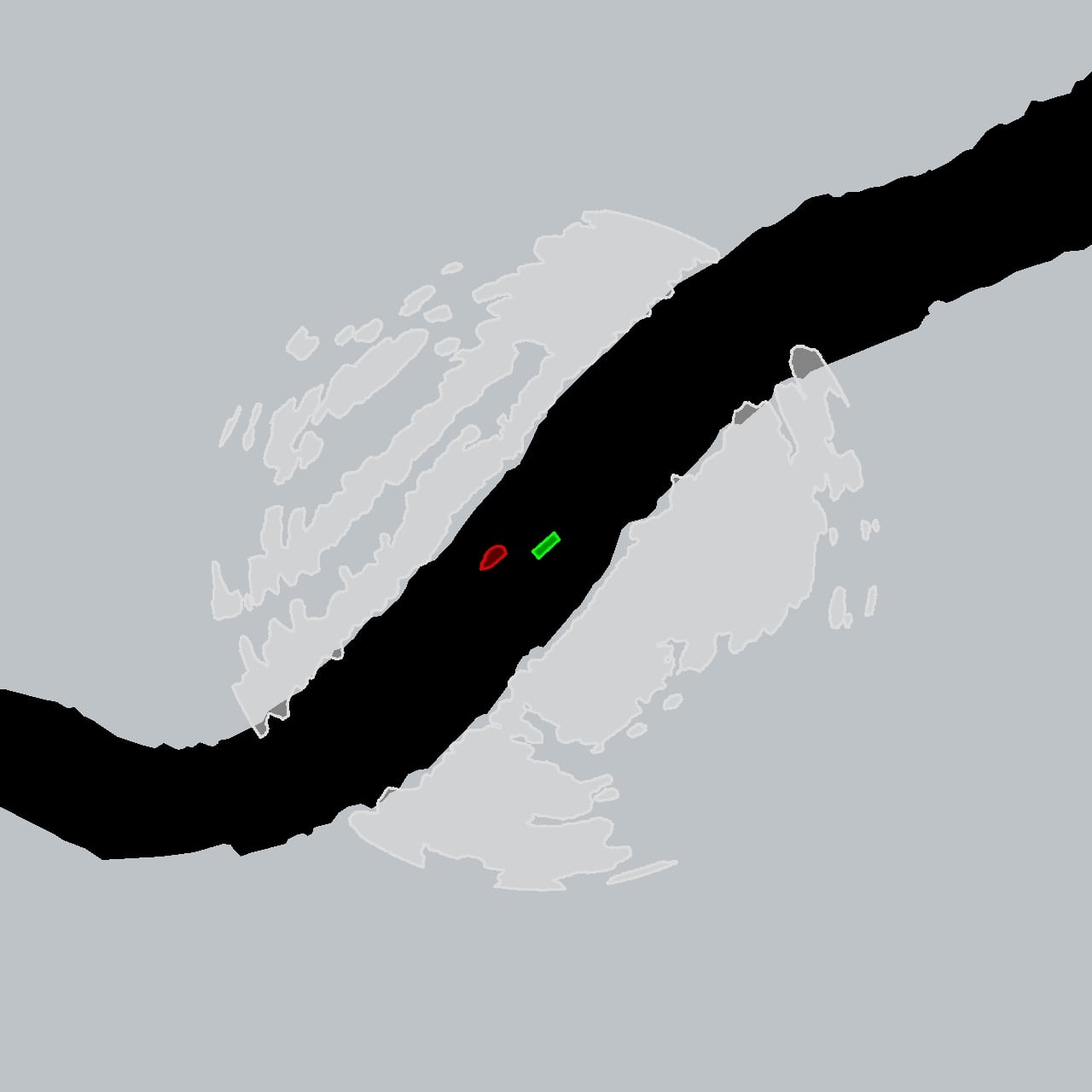}
		\caption{Binary occupancy grid}		
	\end{subfigure}
	\hspace{.2in}
	\begin{subfigure}[h]{3.0in}
		\centering
		\includegraphics[width=3.0in]{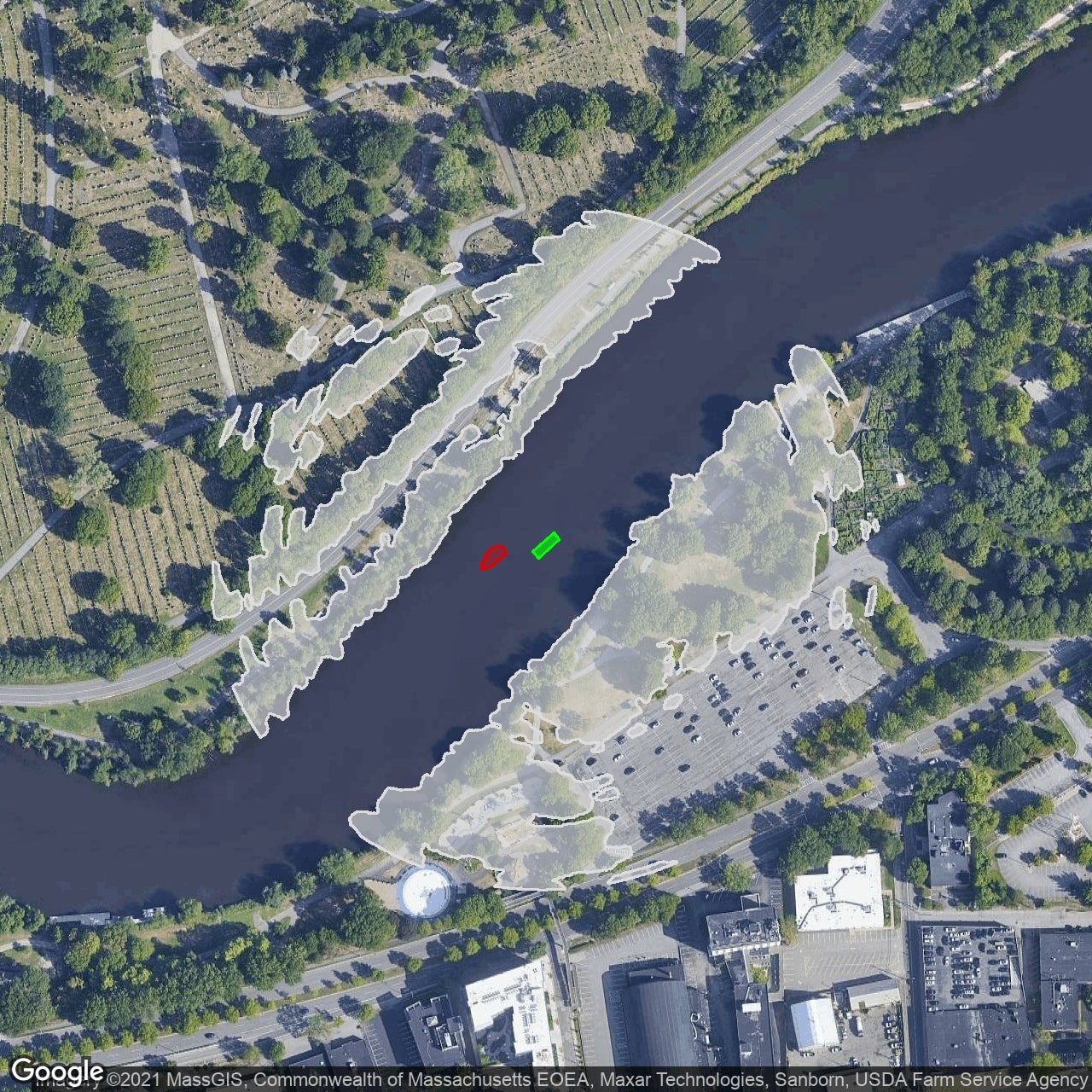}
		\caption{Satellite view}
	\end{subfigure}
	\caption{Processed radar data. The red polygon is the potential target, and the gray polygons on land are considered stationary land objects. The green rectangle is the Jetyak itself. We note that the radar returns often encompass parts of the river which we conservatively classify as land.}
	\label{processed_radar_data}
\end{figure}

For each radar scan, nearby radar intensities are first grouped by connected component labeling. These connected components become separate polygons in the binary occupancy grid map. If the radar polygons are verified to intersect the land polygons that we extracted, we consider them to be stationary land objects and merge these radar polygons with the land area polygons. Otherwise, the radar polygon is considered a potential target in the water. All the potential targets are kept for the further processing. Figure \ref{processed_radar_data}, shows an example with the red target polygons we extracted as a candidate target and the gray polygons on land that we classified as stationary land objects. Our ASV is represented as a green rectangle. We take the center of all the target polygons and track them individually with a Kalman Filter.

\section{Target Tracking}
Although tracking targets by Automatic Identification System (AIS) is available on some marine radars, and used for ASV applications by Kazimierski et al. \cite{Radar_and_Automatic_Identification_System_Track_Fusion_in_an_Electronic_Chart_Display_and_Information_System} and Tetreault et al. \cite{Use_of_the_Automatic_Identification_System_AIS_for_maritime_domain_awareness_MDA}, our tracking algorithm when integrated with an ENC is simpler, open-source and available for any radar platform.

\subsection{Kalman Filter}
A Kalman Filter is implemented to determine a given target's position in $x$ and $y$, and its velocity along $x$ and $y$. In our case, only $x$ and $y$ can be measured for the target, which are derived in the previous step described above. A constant velocity model is applied to track the target's motion.

The state vector is given by
\begin{equation}
x = 
\begin{bmatrix}
p_x \quad p_y \quad v_x \quad v_y
\end{bmatrix}
^T
\end{equation}

$p_x$: distance to the east in UTM

$p_y$: distance to the north in UTM

$v_x$: speed along the x (east) axis

$v_y$: speed along the y (north) axis

The initial state noise covariance matrix $P$ is defined as:

\begin{equation}
P=
\begin{bmatrix}
\sigma_{p_x}^2&0&0&0\\
0&\sigma_{p_y}^2&0&0\\
0&0&\sigma_{v_x}^2&0\\
0&0&0&\sigma_{v_y}^2
\end{bmatrix}
\end{equation}

Prediction stage: We predict the vessel's motion based on a constant velocity model where the process model is given by
\begin{equation}
x^{\prime} = 
Fx + \nu
\end{equation}

Because the target's motion does not always follow a constant velocity model, we add the noise term $\nu$ representing the additional acceleration.

\begin{equation}
\begin{bmatrix}
p_x^{\prime}\\
p_y^{\prime}\\
v_x^{\prime}\\
v_y^{\prime}\\
\end{bmatrix}
=
\begin{bmatrix}
1&0&\Delta t&0\\
0&1&0&\Delta t\\
0&0&1&0\\
0&0&0&1\\
\end{bmatrix}
\begin{bmatrix}
p_x\\
p_y\\
v_x\\
v_y\\
\end{bmatrix}
+
\begin{bmatrix}
\frac{\Delta t^2}{2}\\
\frac{\Delta t^2}{2}\\
\Delta t\\
\Delta t\\
\end{bmatrix}
\begin{bmatrix}
a_x\\
a_y\\
a_x\\
a_y\\
\end{bmatrix}
\end{equation}

We can decompose the noise term $\nu$:
\begin{equation}
\nu = 
\begin{bmatrix}
\frac{\Delta t^2}{2}\\
\frac{\Delta t^2}{2}\\
\Delta t\\
\Delta t\\
\end{bmatrix}
\begin{bmatrix}
a_x\\
a_y\\
a_x\\
a_y\\
\end{bmatrix}
=
\begin{bmatrix}
\frac{\Delta t^2}{2}&0\\
0&\frac{\Delta t^2}{2}\\
\Delta t&0\\
0&\Delta t\\
\end{bmatrix}
\begin{bmatrix}
a_x\\
a_y\\
\end{bmatrix}
=
Ga
\end{equation}

$Q$ is the process noise covariance matrix, which is the expectation of $\nu\nu^T$

\begin{equation}
Q = E[\nu\nu^T] = E[Gaa^TG^T]
\end{equation}

As $G$ does not contain any random variable we can rewrite this as

\begin{equation}
Q = GE[aa^T]G^T = G\begin{bmatrix}
\sigma_{a_x}^2 & \sigma_{a_{xy}}\\
\sigma_{a_{xy}} & \sigma_{a_y}^2\\
\end{bmatrix}
G^T
=
GQ_{\nu}G^T
\end{equation}

As $a_x$ and $a_y$ have no correlation, which means $\sigma_{a_{xy}}=0$, we can simplify  $Q_{\nu}$

\begin{equation}
Q_{\nu} = 
\begin{bmatrix}
\sigma_{a_x}^2 & 0\\
0 & \sigma_{a_y}^2\\
\end{bmatrix}
\end{equation}

Thus, the $Q$ matrix becomes

\begin{equation}
Q = GQ_{\nu}G^T =
\begin{bmatrix}
\frac{\Delta t^4}{4}\sigma_{a_x}^2 & 0 & \frac{\Delta t^3}{2}\sigma_{a_x}^2 & 0\\
0 & \frac{\Delta t^4}{4}\sigma_{a_y}^2 & 0 & \frac{\Delta t^3}{2}\sigma_{a_y}^2\\
\frac{\Delta t^3}{2}\sigma_{a_x}^2 & 0 & \Delta t^2\sigma_{a_x}^2 & 0\\
0 & \frac{\Delta t^3}{2}\sigma_{a_y}^2 & 0 & \Delta t^2\sigma_{a_y}^2\\
\end{bmatrix}
\end{equation}

So that we can compute the prediction state covariance as
$P^{\prime}=FPF^T+Q$

Update stage: We update the vessel's position and velocity based on the radar measurement.

The radar measurement $z$ includes position $p_x$ and $p_y$:
\begin{equation}
z=
\begin{bmatrix}
p_x\\
p_y
\end{bmatrix}
\end{equation}

The measurement noise covariance matrix $R$ can be defined as:

\begin{equation}
R=
\begin{bmatrix}
\sigma_{p_x}^2&0\\
0&\sigma_{p_y}^2
\end{bmatrix}
\end{equation}

We also define matrix $H$  which is used to transform state $x$ to measurement $z$:

\begin{equation}
\begin{bmatrix}
p_x\\
p_y
\end{bmatrix}
=
H
\begin{bmatrix}
p_x^{\prime}\\
p_y^{\prime}\\
v_x^{\prime}\\
v_y^{\prime}
\end{bmatrix}
\end{equation}

Where
\begin{equation}
H=
\begin{bmatrix}
1&0&0&0\\
0&1&0&0
\end{bmatrix}
\end{equation}

So that we can eventually get 

The measurement residual:$y=z-Hx^{\prime}$

Residual covariance: $S=HP^{\prime}H^T+R$

Kalman gain: $K=P^{\prime}H^TS^{-1}$

Updated state estimation: $x=x^{\prime}+Ky$

Updated state covariance: $P=(I-KH)P^{\prime}$

\subsection{Data Association}
\begin{figure} [h]
    \centering
    \includegraphics[height=2.3in]{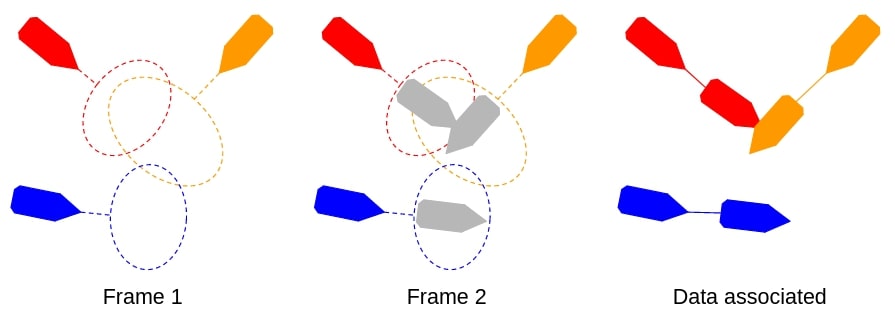}
    \caption{Data association. The Kalman Filter calculates each target's predicted state estimation, including a position with covariance, as illustrated by the ellipses in the figure. We maximize the sum of all the likelihoods of all the associations between two frames. The associations with the highest probability are selected.}
    \label{data_association}
\end{figure}

We run a separate Kalman Filter for each target that is identified in the radar scans. 
We assume that any object creates only one single measurement (one polygon in the binary occupancy grid), and one measurement can be created only by a single object. To disambiguate multiple targets we use a data association algorithm. The Kalman Filter calculates a predicted state estimation for each target for the next time step, including a predicted position and its covariance. We use the elliptical shape of the covariance as a gate. If a target in a succeeding frame is in the gate associated with a target's predicted state in the current frame, then these two targets are considered to be the same. If there are multiple targets in the same gate in the next frame, we compute the likelihood of every target in this gate associated with the corresponding target in the original frame. The probability density function given by the Kalman Filter can be applied to get the probability, which corresponds to the likelihood in this case can be use for data association. We aim to maximize the sum of all the likelihoods for all the associations. The associations with the highest probability are selected. In practise this method works extremely well as illustrated in Figure \ref{data_association}. 

If a new target shows up in a frame, then it would be associated with a "null" target in the previous frame, which leads to a likelihood of zero, so that the new target can be recognized and initialized. If a target disappears in a succeeding frame, the target in the current frame would be associated with a "null" target.

Associations of targets are checked in two ways to prevent highly improbable results. As a first check, we use the maximum distance we believe any given target could travel in the time between measurements, $range=v\cdot t_{meas}$. In our case, since $t_{meas}\approx 2.5s$ for each radar scan and we are highly unlikely to encounter any vessels on the Charles with a $v>10m/s$, we use $range=25m$. In environments where we do expect to see faster vessels, $range$ can be increased accordingly. Another check is implemented by restricting the size of the gate. The likelihood computed by the probability density function (pdf) has a threshold that restricts the size of the covariance ellipse. In our case, we use a threshold of $10^{-6}$.

\begin{figure} [h]
    \centering
    \includegraphics[height=3.5in]{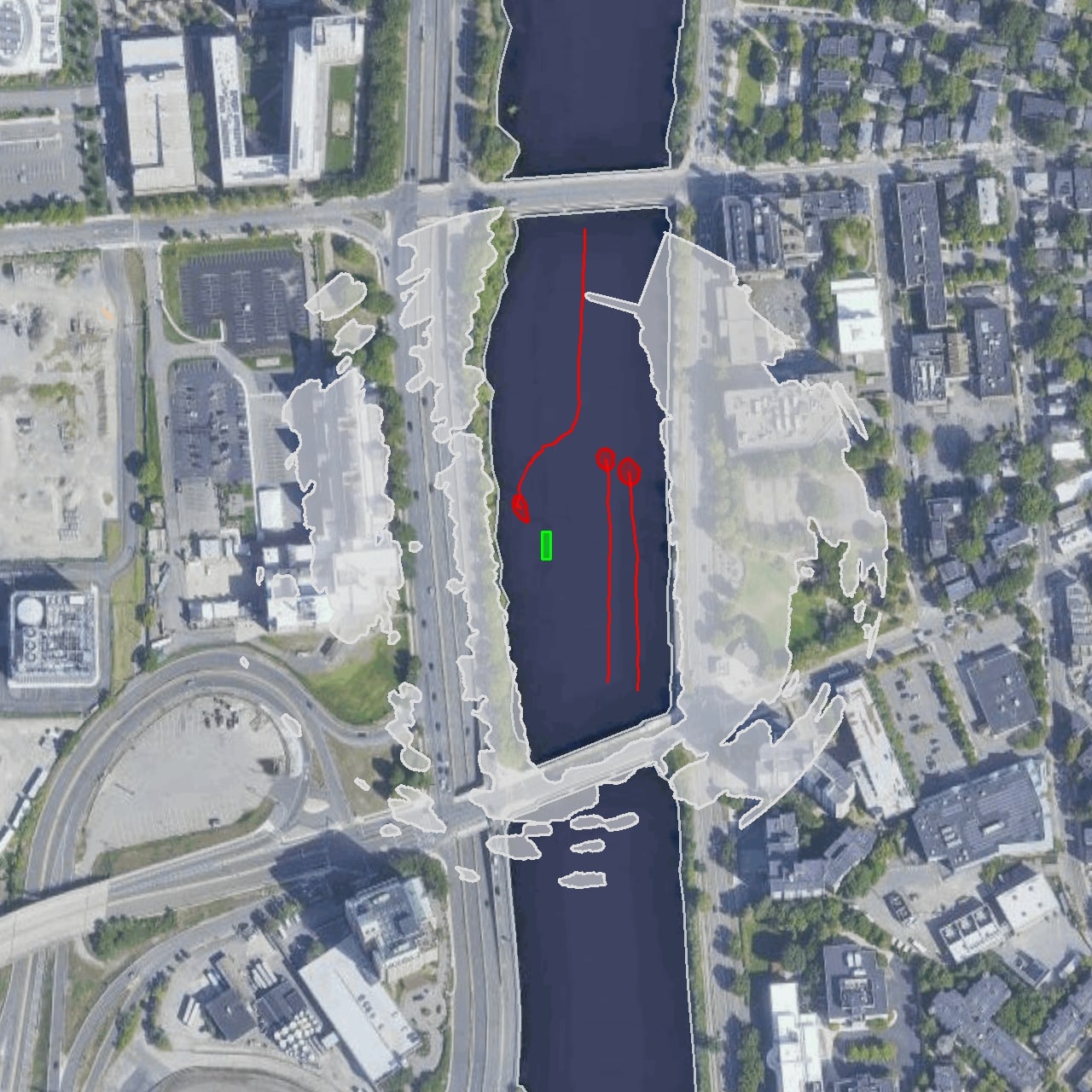}
    \caption{The Result of Data association. There are three target vessels near the Jetyak. We detect each target separately and associate their trajectories across multiple frames.}
    \label{data_association_result}
\end{figure}

Figure \ref{data_association_result} shows the results of our data association algorithm for multiple targets across multiple frames. The red lines in the figure are the trajectories of those targets.

After the data association, measurements can be input to the Kalman Filter to run the update functions. Then the posterior probability of each target is retrieved as the current state of the Kalman Filter, including UTM coordinates $p_x$ and $p_y$, velocity along each axis $v_x$ and $v_y$, and corresponding state covariance $P$.

Sometimes the target disappears for a single frame but then reappears. Typically, this happens when the target is too close to the coastline and gets lumped as a land object. Two frames are cached to avoid losing track of such targets in such a situation. As there is no measurement in the frame where the target disappears, only the Kalman Filter's prediction step is run.

\section{COLREGs Projection}
In order to avoid other target vessels in accordance with COLREGs, a virtual forbidden polygon is applied to each vessel in accordance with the active COLREGs rule. 

To do this, first, a check is performed to see whether the target vessel is in front of the Jetyak. Only if the target vessel is in front of our ASV do we take action to enforce COLREGs on the Jetyak.

\begin{figure}
	\centering
	\begin{subfigure}[b]{2.5in}
		\centering
		\includegraphics[width=2.5in]{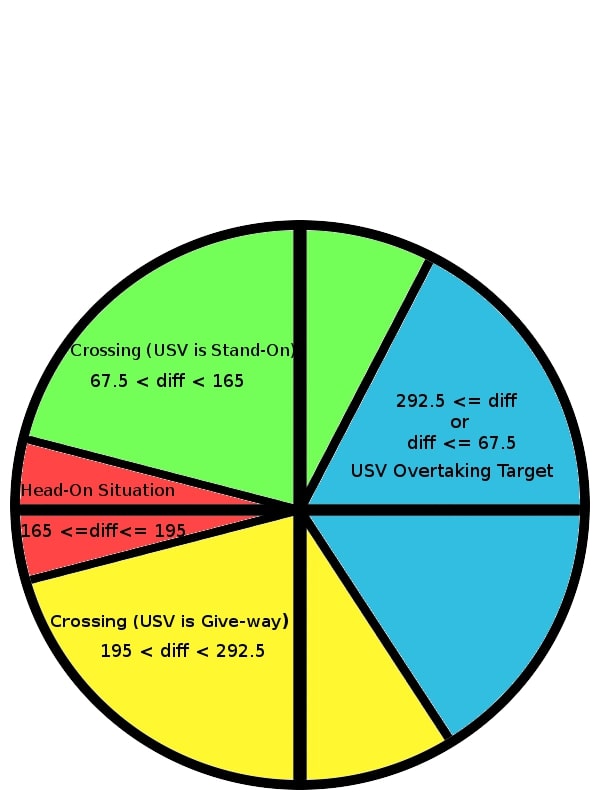}
		\caption{COLREGs situations}
		\label{COLREGs_situations}
	\end{subfigure}
	\hspace{.2in}
	\begin{subfigure}[b]{3.5in}
		\centering
		\includegraphics[width=3.5in]{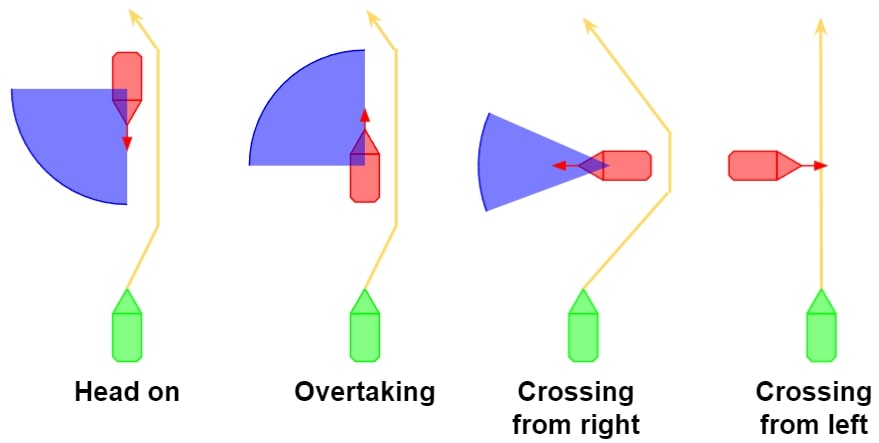}
		\caption{COLREGs projections. We pad the additional blue projections onto the target polygons according to the COLREGs situation, ensuring that the Jetyak modifies its behavior to follow COLREGs.}
		\label{COLREGs_projection}
	\end{subfigure}
\end{figure}


\begin{figure}
	\centering
	\begin{subfigure}[h]{3.0in}
		\centering
		\includegraphics[width=3.0in]{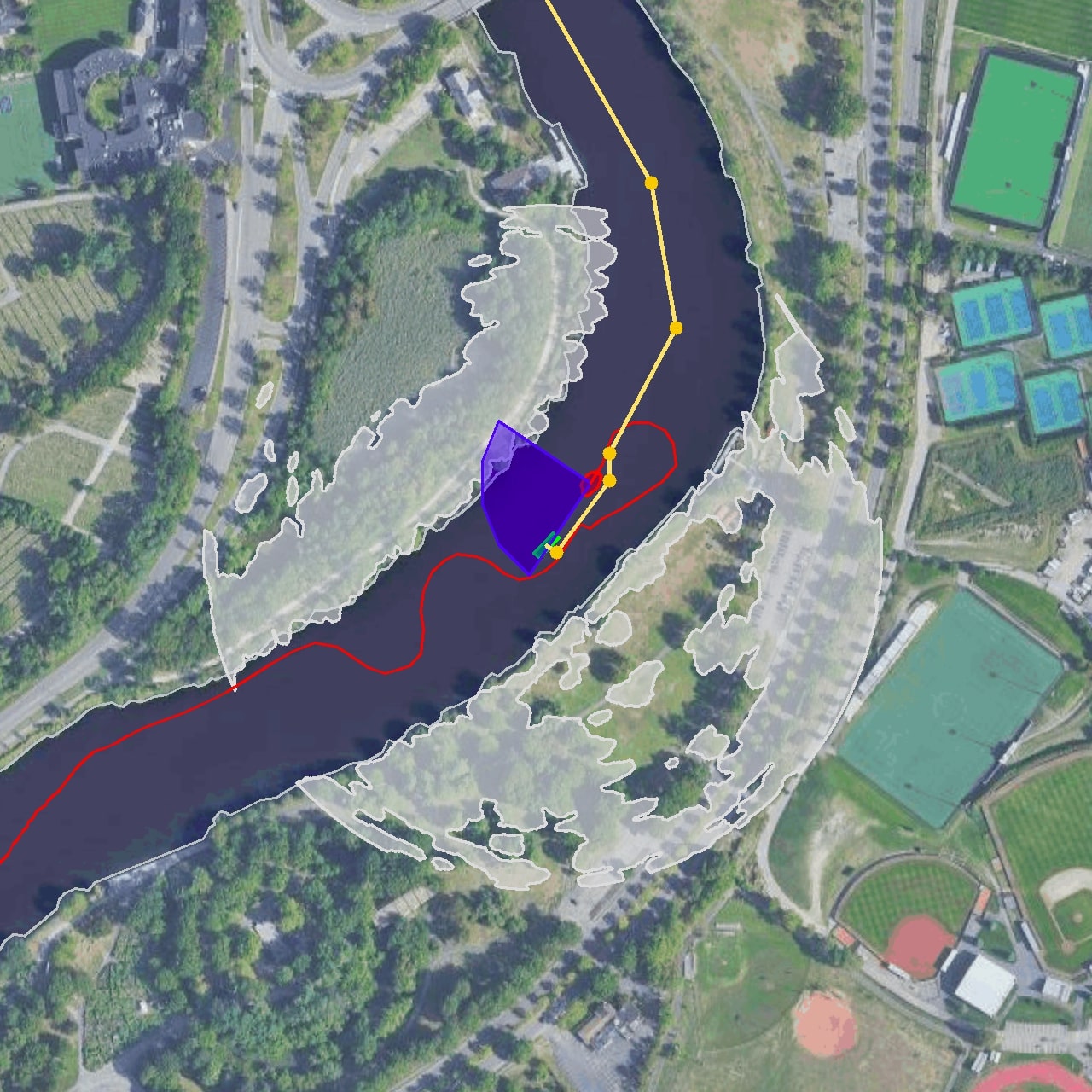}
		\caption{Head on: COLREGs require the Jetyak to pass on the right side of the target vessel.}		
	\end{subfigure}
	\hspace{.2in}
	\begin{subfigure}[h]{3.0in}
		\centering
		\includegraphics[width=3.0in]{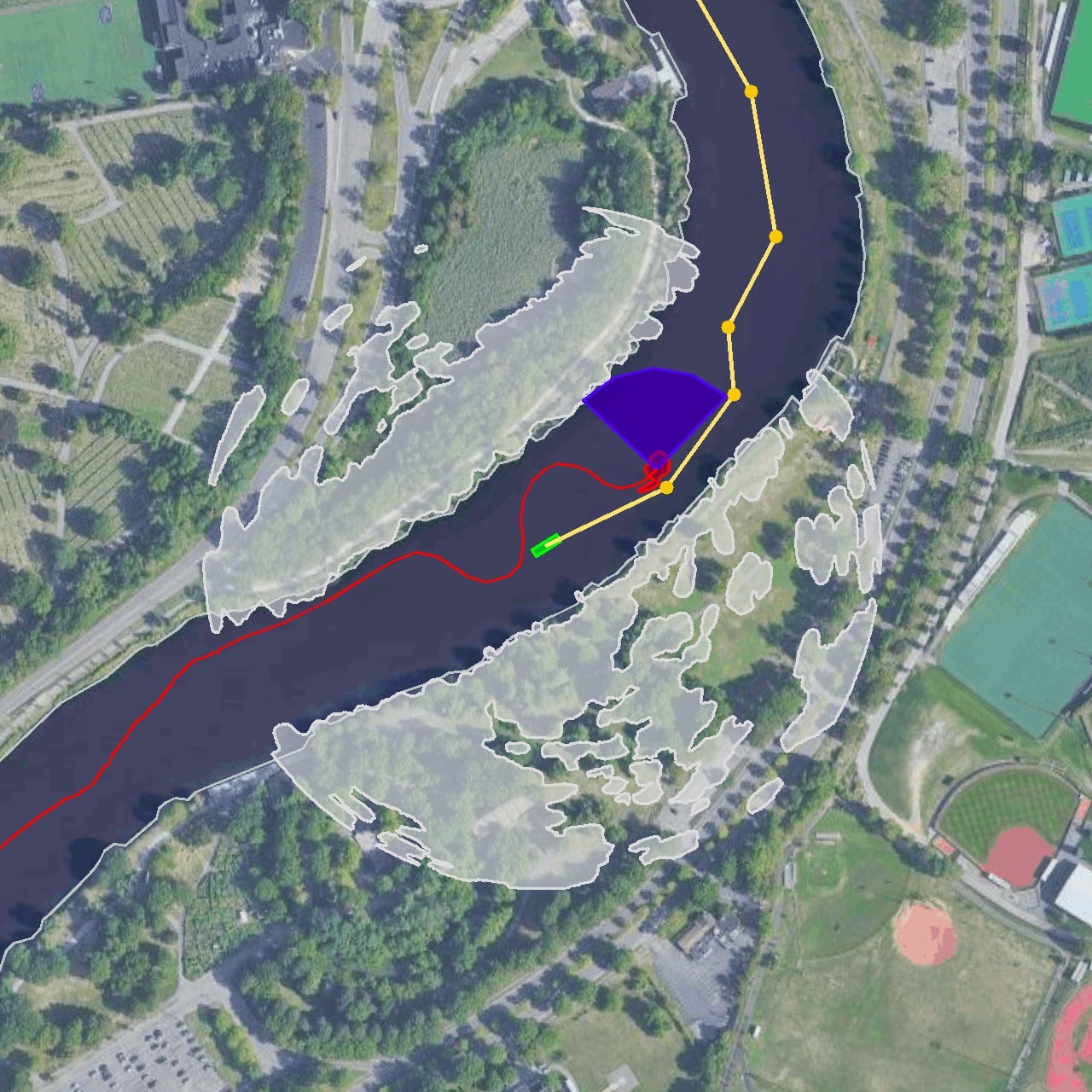}
		\caption{Overtaking: COLREGs require the Jetyak to overtake the target vessel from the right side.}
	\end{subfigure}
	\\\vspace{.2in}
	\begin{subfigure}[h]{3.0in}
		\centering
		\includegraphics[width=3.0in]{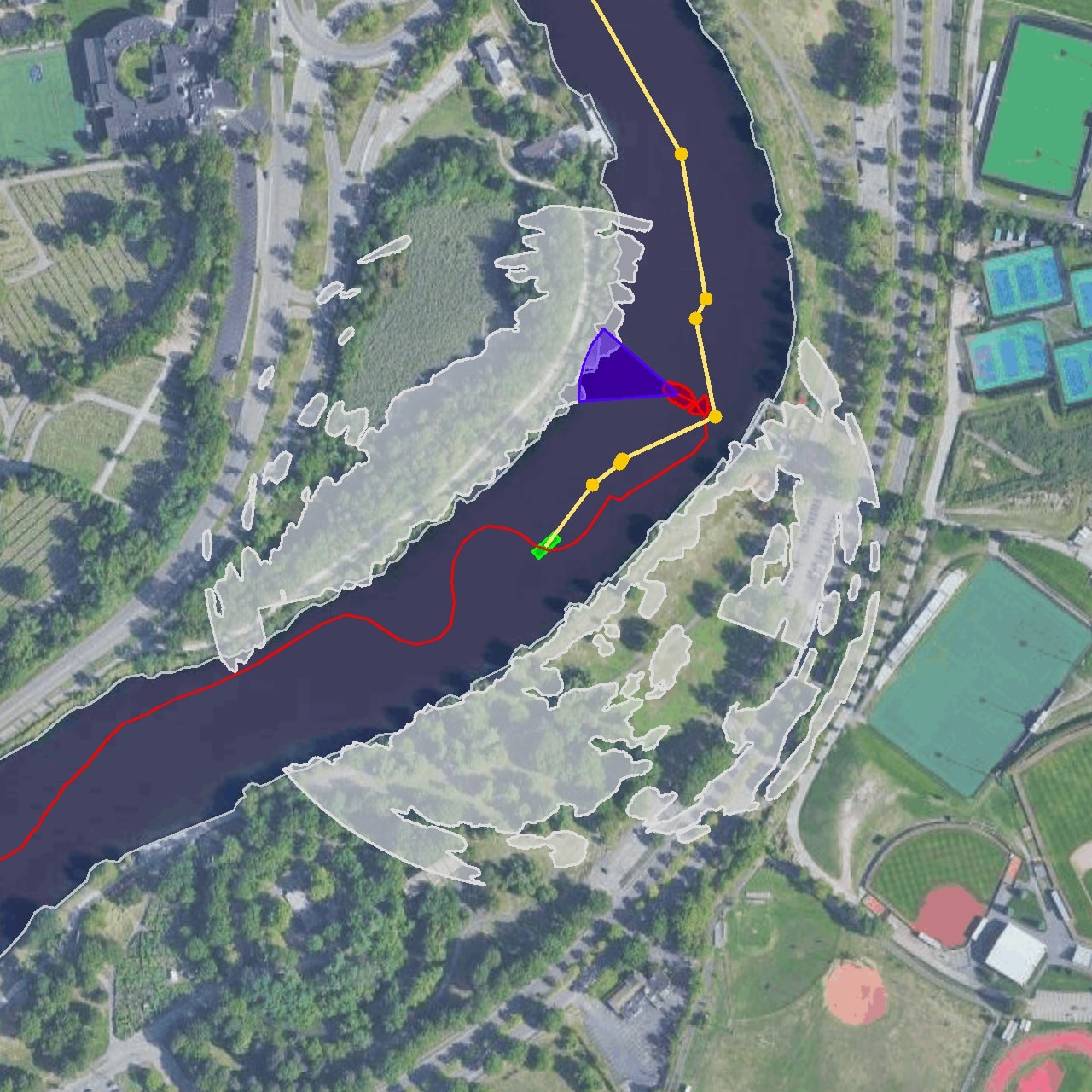}
		\caption{Crossing from the right: COLREGs require that the Jetyak not block the target vessel when it is crossing from the right to the left.}		
	\end{subfigure}
	\hspace{.2in}
	\begin{subfigure}[h]{3.0in}
		\centering
		\includegraphics[width=3.0in]{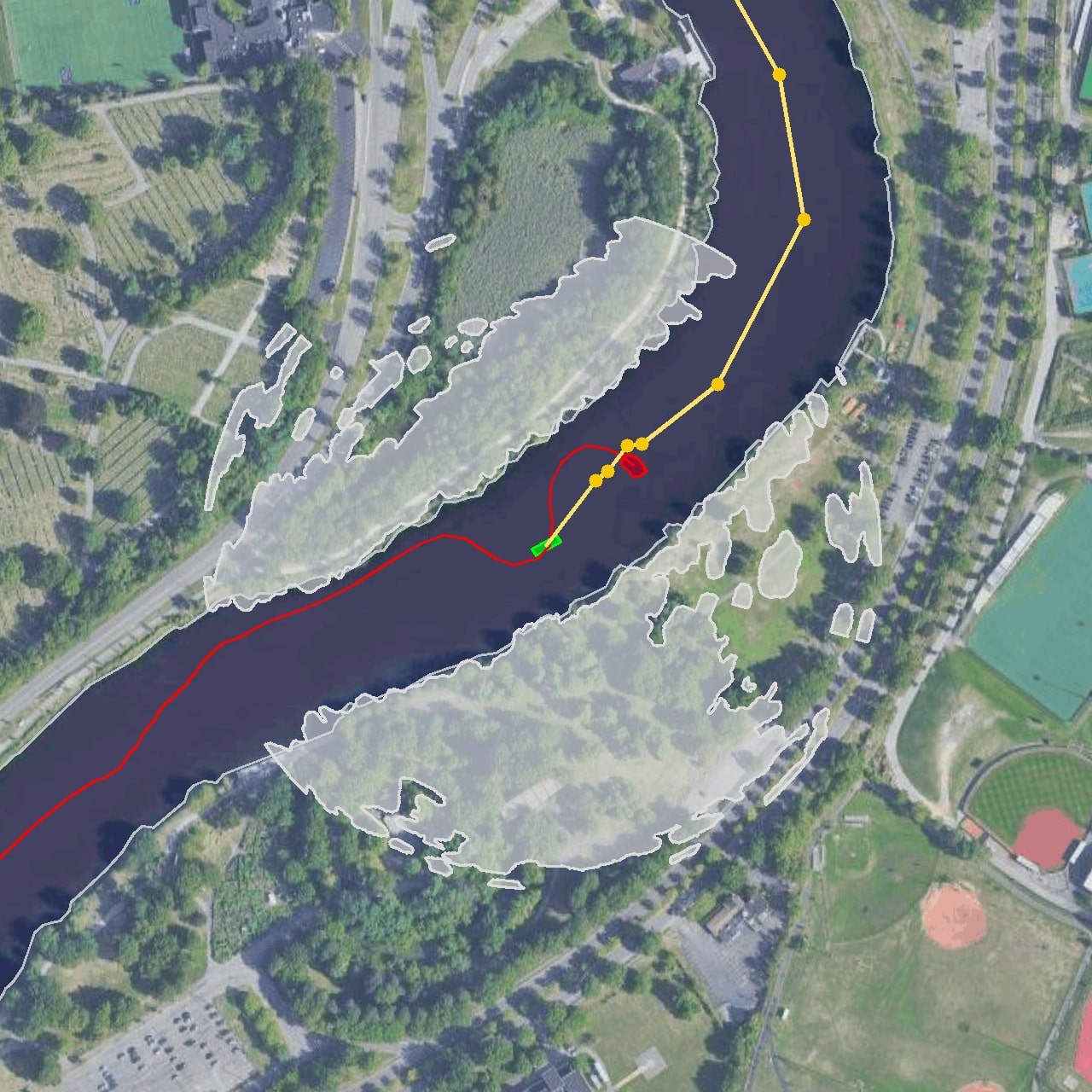}
		\caption{Crossing from the left: COLREGs requires the Jetyak to hold course and not give way to the target vessel when the target is crossing from the left to the right.}
	\end{subfigure}
	\caption{Applying different projections according to the different COLREGs situations.}
	\label{COLREGs_projections_results}
\end{figure}

If the target vessel is determined to be in front of us, we need to decide what actions have to be taken according to the COLREGs situation. The COLREGs situation is determined by computing the difference between the Jetyak's heading $\theta_a$ and the target's heading $\theta_b$ where $diff=\theta_a-\theta_b$. Figure \ref{COLREGs_situations} shows the relationship between the heading difference and how that translates to vessel interactions under COLREGs. After considering the difference, the applicable COLREGs projection is applied as an forbidden area in the binary occupancy grid. Figure \ref{COLREGs_projection} shows the different COLREGs projections corresponding to the COLREGs situations. We can see the yellow line (planed path) successfully avoids the target vessel while following COLREGs in each case. Figure \ref{COLREGs_projections_results} shows the results of applying COLREGs velocity obstacles in different situations while running our algorithm for real world situations associated with our test deployments. 

\section{Path planning algorithm}
Finally, we can account for the COLREGs projections by using a reactive motion planning algorithm based on elements of the one proposed by Kuwata et al. \cite{Safe_Maritime_Autonomous_Navigation_With_COLREGS_Using_Velocity_Obstacles}. Rather than evaluating an optimal trajectory for the USV as Kuwata does, the velocity obstacles are represented as polygons in the binary occupancy grid of the environment. When the risk of collision is deemed to exist for the Jetyak's current set of waypoints, a new set of waypoints is created through the use of Visibility Graph Inspired Path Planning (VGIPP). These waypoints modify the global path for the Jetyak and are executed to avoid collisions related to COLREGs. Using VGIPP is computationally very efficient for our application.

VGIPP is based on the binary occupancy grid map. All the stationary and dynamic obstacles with their COLREGs projections are represented as occupied polygons on the map. The part of the grid that is not occupied is considered safe for traversal.

First, we pre-define a global path for the entire trip. As our river transit involves sections that are relatively narrow in parts, we choose a path that keeps us in the center of the river. This is not the shortest path, but avoids getting the vessel very close to shore. 

\begin{figure} [h]
    \centering
    \includegraphics[height=2.0in]{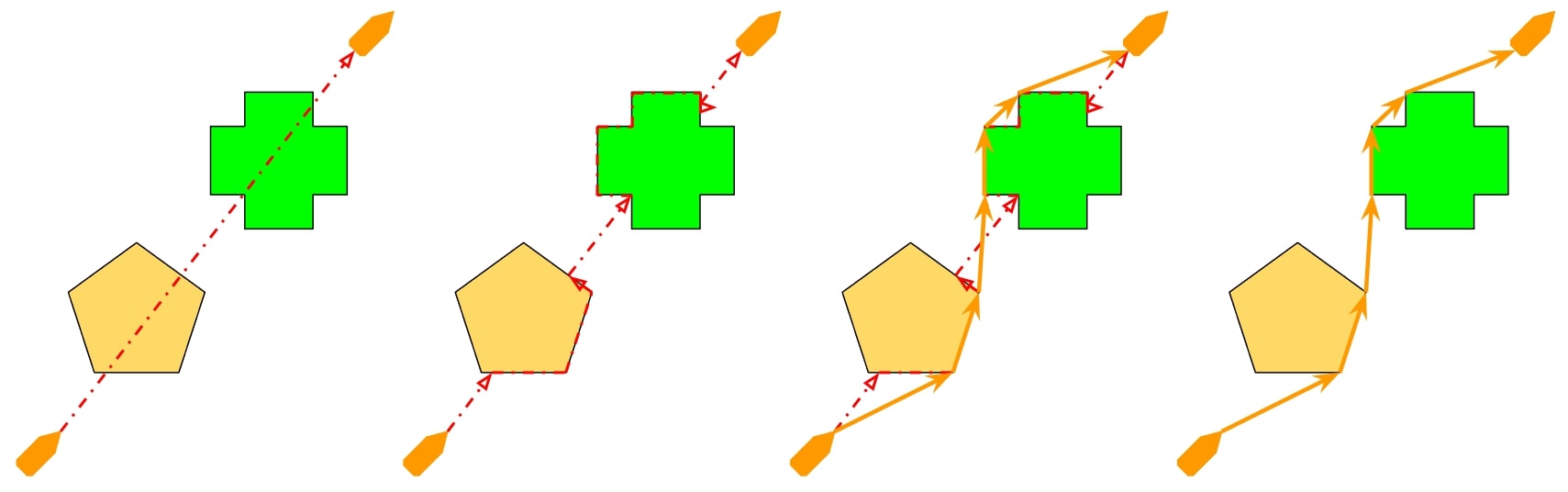}
    \caption{The Path Planning algorithm. We predefine a safe path (step 1) independent of any obstacles. We then set the shorter contour of obstacle polygons as the temporary safe path (step 2). We continue to refine the path by checking each point from the end point to the start point, to see whether this point can directly connect to the start point. If a shorter path is detected, we take that as the new path. After checking all the points after the start point, we repeat the procedure with the second point as the start point, and check all the points after it. After all the points are checked in this way, the final path is selected (steps 3 and 4).}
    \label{path_planning_algorithm}
\end{figure}

\begin{figure}
	\centering
	\begin{subfigure}[h]{3.0in}
		\centering
		\includegraphics[width=3.0in]{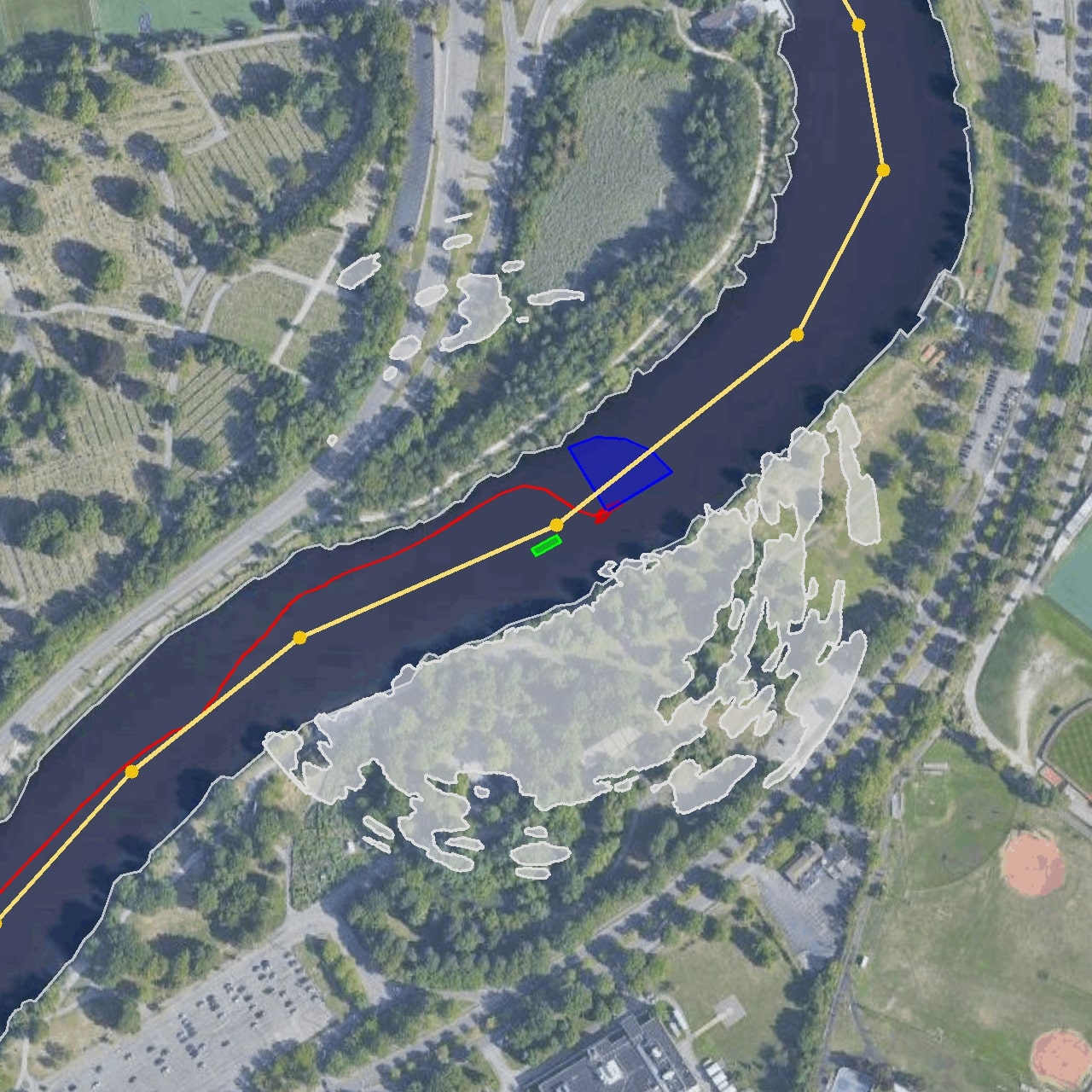}
		\caption{We predefine a safe path in the middle of the river.}		
	\end{subfigure}
	\hspace{.2in}
	\begin{subfigure}[h]{3.0in}
		\centering
		\includegraphics[width=3.0in]{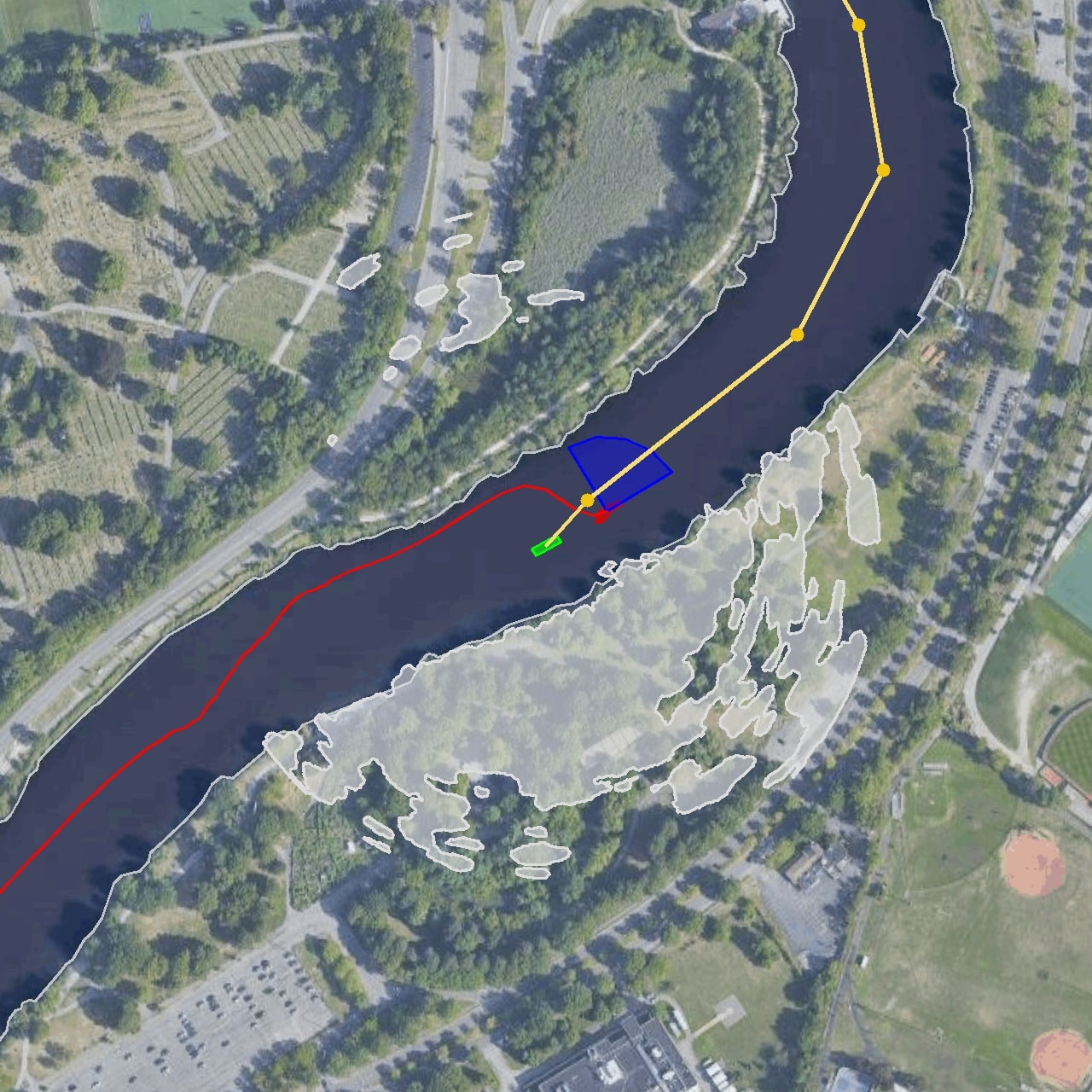}
		\caption{We check to see if the path intersects with any obstacles between the start point (Jetyak) and the end point (destination).}
	\end{subfigure}
	\\\vspace{.2in}
	\begin{subfigure}[h]{3.0in}
		\centering
		\includegraphics[width=3.0in]{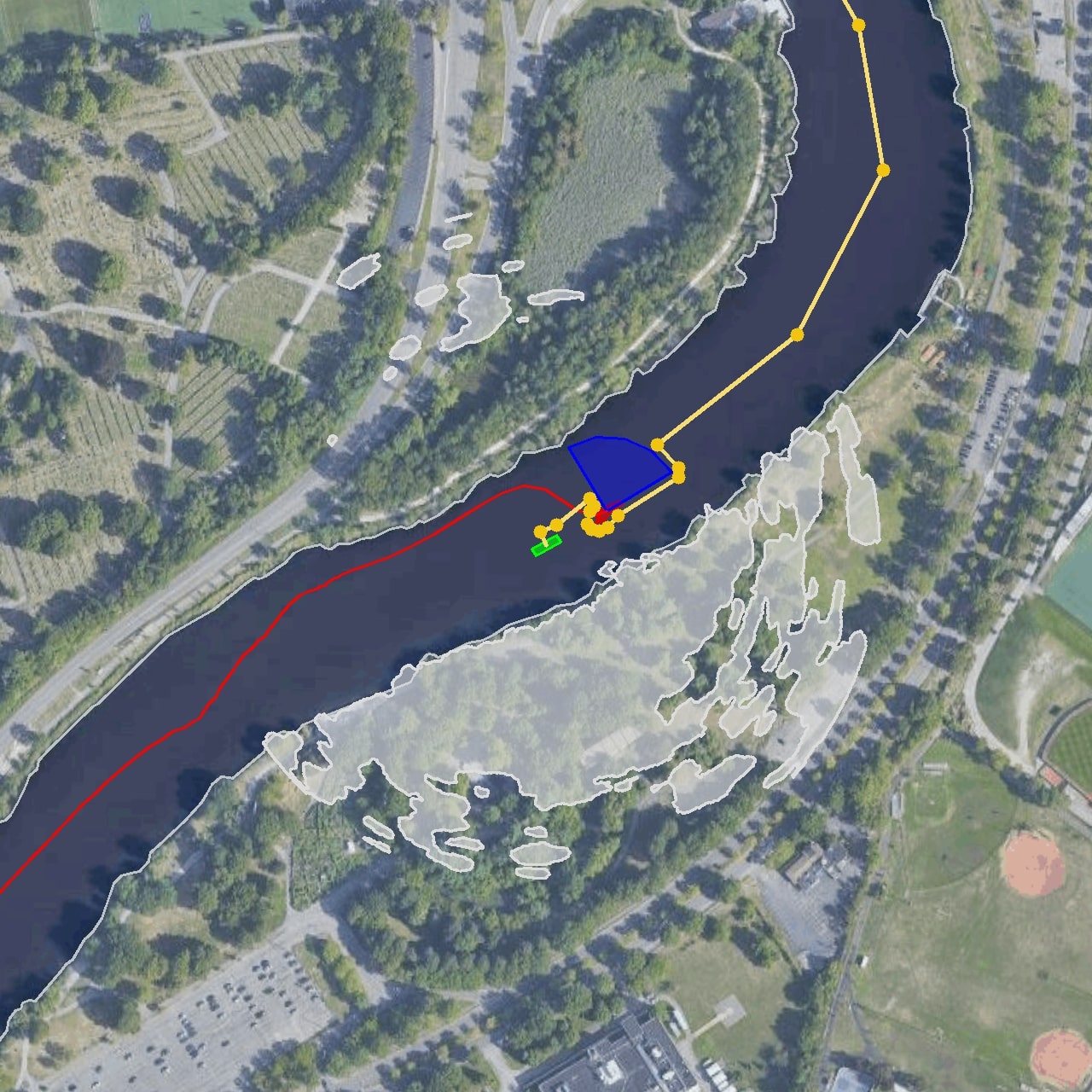}
		\caption{We then get the shorter contour of obstacle polygons as the temporary safe path.}		
	\end{subfigure}
	\hspace{.2in}
	\begin{subfigure}[h]{3.0in}
		\centering
		\includegraphics[width=3.0in]{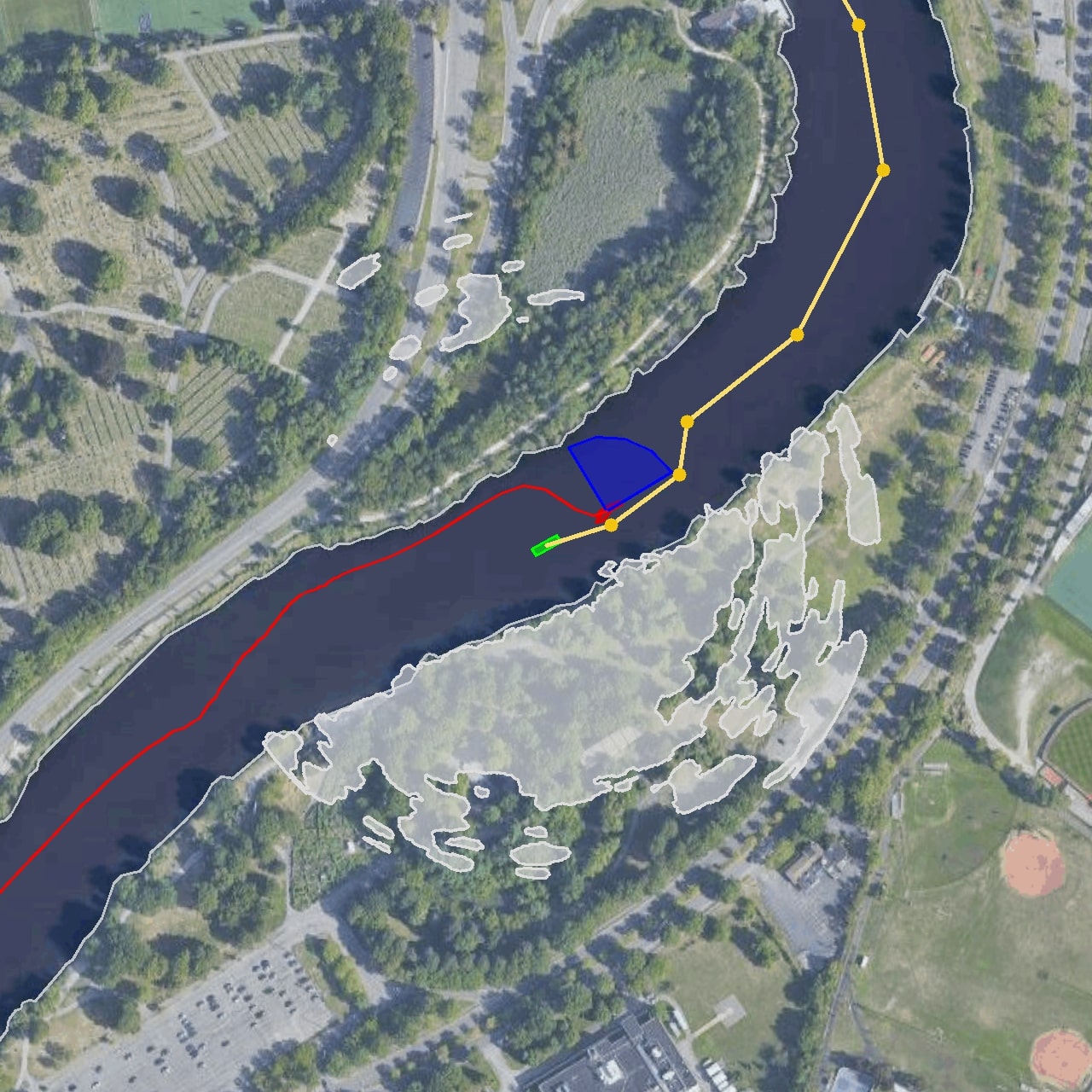}
		\caption{Finally, we optimize the temporary path by checking to see whether there is a shorter path in the route.}
	\end{subfigure}
	\caption{Path planning algorithm results.}
	\label{path_planning_algorithm_result}
\end{figure}

While transiting down the river we are continually running VGIPP as illustrated in Figure \ref{path_planning_algorithm}. The predefined path may intersect with some obstacle polygons, and we find the contours of these polygons. These contours can, in turn, be split into two sets of polylines. We use the shorter polyline as our temporary optimal path. We then check each point from the end point to the start point, to see whether the point under consideration can directly connect to the start point. If a shorter path is detected, we use the path as our new optimal path. After checking all the points after the start point, we repeat the procedure using the second point as our start point, and checking all the points after it. After all the points have been checked in this way, the final path emerges. Figure \ref{path_planning_algorithm_result} shows how VGIPP is implemented.

If the Jetyak is already in an unsafe area, such as the COLREGs projection of a target, it tries to escape that area first. The algorithm finds the nearest safe point and creates a path to it, and then continues to find a safe path to the destination.

This algorithm is not guaranteed to yield an optimal result, but it is very efficient in the binary occupancy grid. In our case, the size of the binary occupancy grid map is $4000^2$. A traditional algorithm could take orders of magnitude longer to calculate the shortest path in such a huge and detailed map for each scan. The VGIPP works efficiently in this case, which takes only around 0.2s to get the result with Python. In most cases, VGIPP does yield the optimal path. In complex environments, it also guarantees that the safe path can be detected if one exists. We note that if no acceptable path is found, the Jetyak can slow down and come to a complete stop.

\section{Conclusions and Future Work}
In this paper we have presented an end-to-end system that is capable of navigating in constrained waters while respecting the constraints associated with COLREGs. Our minimal sensor suite consists of GPS, IMU and Radar measurements. 

Currently, we can collect data from radar, GPS, IMU, and compass sensors with ROS and replay the data to calculate obstacle-free waypoints that comply with COLREGs. While those waypoints are not executed live on the Jetyak due to safety and permitting concerns, they are calculated in real-time with real data and can be used to avoid collision with both dynamic and static obstacles in a complicated environment. The processing time of the whole algorithm in Python is around 0.3s per radar scan on an i7-10750H CPU (2.60GHz with 6 cores). 



We continue to work on extending our algorithm to include sensors such as cameras and lidars to allow us to work in more diverse areas, such as coming up and crossing under a bridge where our algorithm would fail due to shortcomings in the radar. We would also like to extend our work to areas that are GPS denied such as rivers that are narrow and covered by trees. 

\bibliographystyle{apalike}
\bibliography{frExampleRefs}

\begin{thebibliography}{}

\bibitem[Benjamin and Curcio,
  2004]{COLREGS_based_navigation_of_autonomous_marine_vehicles}
Benjamin, M. and Curcio, J. (2004).
\newblock Colregs-based navigation of autonomous marine vehicles.
\newblock In {\em 2004 IEEE/OES Autonomous Underwater Vehicles (IEEE Cat.
  No.04CH37578)}, pages 32--39.

\bibitem[Benjamin et~al.,
  2006]{A_method_for_protocol_based_collision_avoidance_between_autonomous_marine_surface_craft}
Benjamin, M., Leonard, J., Curcio, J., and Newman, P. (2006).
\newblock A method for protocol-based collision avoidance between autonomous
  marine surface craft.
\newblock {\em J. Field Robotics}, 23:333--346.

\bibitem[Blaich et~al.,
  2015]{Mission_integrated_collision_avoidance_for_USVs_using_laser_range_finder}
Blaich, M., Köhler, S., Schuster, M., Schuchhardt, T., Reuter, J., and Tietz,
  T. (2015).
\newblock Mission integrated collision avoidance for usvs using laser range
  finder.
\newblock In {\em OCEANS 2015 - Genova}, pages 1--6.

\bibitem[br24, ]{br24}
br24.
\newblock https://github.com/juancamilog/br24.

\bibitem[Dubey and Louis,
  2021]{VORRT_COLREGs_A_Hybrid_Velocity_Obstacles_and_RRT_Based_COLREGs_Compliant_Path_Planner_for_Autonomous_Surface_Vessels}
Dubey, R. and Louis, S. (2021).
\newblock Vorrt-colregs: A hybrid velocity obstacles and rrt based
  colregs-compliant path planner for autonomous surface vessels.

\bibitem[(IMO),
  1972]{COLREG_Convention_on_the_International_Regulations_for_Preventing_Collisions_at_Sea}
(IMO), I. M.~O. (1972).
\newblock {\em COLREG: Convention on the International Regulations for
  Preventing Collisions at Sea}.
\newblock London: International Maritime Organization, 2003.

\bibitem[Johansen et~al.,
  2016]{Ship_Collision_Avoidance_and_COLREGS_Compliance_Using_Simulation_Based_Control_Behavior_Selection_With_Predictive_Hazard_Assessment}
Johansen, T.~A., Perez, T., and Cristofaro (2016).
\newblock Ship collision avoidance and colregs compliance using
  simulation-based control behavior selection with predictive hazard
  assessment.
\newblock {\em IEEE Transactions on Intelligent Transportation Systems},
  17:3407--3422.

\bibitem[Kazimierski and Stateczny,
  2015]{Radar_and_Automatic_Identification_System_Track_Fusion_in_an_Electronic_Chart_Display_and_Information_System}
Kazimierski, W. and Stateczny, A. (2015).
\newblock Radar and automatic identification system track fusion in an
  electronic chart display and information system.
\newblock {\em Journal of Navigation}, pages 1--14.

\bibitem[Kimball et~al.,
  2014]{The_WHOI_Jetyak_An_autonomous_surface_vehicle_for_oceanographic_research_in_shallow_or_dangerous_waters}
Kimball, P., Bailey, J., Das, S., Geyer, R., Harrison, T., Kunz, C., Manganini,
  K., Mankoff, K., Samuelson, K., Sayre-McCord, T., Straneo, F., Traykovski,
  P., and Singh, H. (2014).
\newblock The whoi jetyak: An autonomous surface vehicle for oceanographic
  research in shallow or dangerous waters.
\newblock In {\em 2014 IEEE/OES Autonomous Underwater Vehicles (AUV)}, pages
  1--7.

\bibitem[Kuwata et~al.,
  2014]{Safe_Maritime_Autonomous_Navigation_With_COLREGS_Using_Velocity_Obstacles}
Kuwata, Y., Wolf, M.~T., Zarzhitsky, D., and Huntsberger, T.~L. (2014).
\newblock Safe maritime autonomous navigation with colregs, using velocity
  obstacles.
\newblock {\em IEEE Journal of Oceanic Engineering}, 39:110--119.

\bibitem[Liu and Bucknall,
  2015]{Path_planning_algorithm_for_unmanned_surface_vehicle_formations_in_a_practical_maritime_environment}
Liu, Y. and Bucknall, R. (2015).
\newblock Path planning algorithm for unmanned surface vehicle formations in a
  practical maritime environment.
\newblock {\em Ocean Engineering}, 97.

\bibitem[mavlink, ]{mavlink}
mavlink.
\newblock https://github.com/mavlink/mavlink.

\bibitem[mavros, ]{mavros}
mavros.
\newblock https://github.com/mavlink/mavros.

\bibitem[Naeem et~al.,
  2012]{COLREGs_based_collision_avoidance_strategies_for_unmanned_surface_vehicles}
Naeem, W., Irwin, G., and Yang, A. (2012).
\newblock Colregs-based collision avoidance strategies for unmanned surface
  vehicles.
\newblock {\em Mechatronics}, 22.

\bibitem[QGIS, ]{QGIS}
QGIS.
\newblock https://qgis.org/en/site.

\bibitem[Schuster et~al.,
  2014]{Collision_Avoidance_for_Vessels_using_a_Low_Cost_Radar_Sensor}
Schuster, M., Blaich, M., and Reuter, J. (2014).
\newblock Collision avoidance for vessels using a low-cost radar sensor.
\newblock {\em IFAC Proceedings Volumes}, 47:9673--9678.

\bibitem[Tetreault,
  2005]{Use_of_the_Automatic_Identification_System_AIS_for_maritime_domain_awareness_MDA}
Tetreault, B. (2005).
\newblock Use of the automatic identification system (ais) for maritime domain
  awareness (mda).
\newblock In {\em Proceedings of OCEANS 2005 MTS/IEEE}, pages 1590--1594 Vol.
  2.

\bibitem[vectornav, ]{vectornav}
vectornav.
\newblock https://github.com/dawonn/vectornav.

\bibitem[Zhang et~al.,
  2015]{A_distributed_anti_collision_decision_support_formulation_in_multi_ship_encounter_situations_under_COLREGs}
Zhang, J., Zhang, D., Yan, X., Haugen, S., and Guedes~Soares, C. (2015).
\newblock A distributed anti-collision decision support formulation in
  multi-ship encounter situations under colregs.
\newblock {\em Ocean Engineering}, 105:336--348.

\end{thebibliography}

\end{document}